\documentclass[lettersize,journal]{IEEEtran}
\usepackage{amsmath,amsfonts}
\usepackage{algorithmic}
\usepackage{algorithm}
\usepackage{array}
\usepackage{textcomp}
\usepackage{stfloats}
\usepackage{url}
\usepackage{verbatim}
\usepackage{graphicx}
\usepackage{cite}
\usepackage{times}
\usepackage{epsfig}
\usepackage{amsmath}
\usepackage{amssymb}
\usepackage{tabularx}
\usepackage{makecell}
\usepackage{pifont}
\usepackage{booktabs}
\usepackage{multirow}
\usepackage{subcaption}
\newcommand{\etal}{\textit{et al}.}

\newcommand{\eg}{\textit{e}.\textit{g}.}
\usepackage[pagebackref=true,breaklinks=true,letterpaper=true,colorlinks,bookmarks=false]{hyperref}
\usepackage{cite}

\begin{document}

\title{EgoVSR: Towards High-Quality Egocentric Video Super-Resolution}

\author{Yichen Chi, Junhao Gu, Jiamiao Zhang, Wenming Yang, \IEEEmembership{Senior Member, IEEE}, Yapeng Tian, \IEEEmembership{Member, IEEE}
\thanks{Y. Chi, J. Gu, J. Zhang and W. Yang are with the Shenzhen International Graduate School, Tsinghua University, Shenzhen 518055, China (Email: chiyc21@mails.tsinghua.edu.cn; gjh21@mails.tsinghua.edu.cn; zjm21@mails.tsinghua.edu.cn; yangelwm@163.com), W. Yang is the corresponding author.}
\thanks{Y. Tian is with the Department of Computer Science, The University of Texas at Dallas, Richardson, TX 75080, USA (E-mail: yapeng.tian@utdallas.edu)}.}

\markboth{Journal of \LaTeX\ Class Files,~Vol.~14, No.~8, August~2021}%
{Shell \MakeLowercase{\textit{et al.}}: A Sample Article Using IEEEtran.cls for IEEE Journals}

\maketitle   

\begin{figure*}
   \captionsetup{type=figure}
	\begin{subfigure}[c]{0.31\textwidth}
		\centering
		\includegraphics[width=\textwidth]{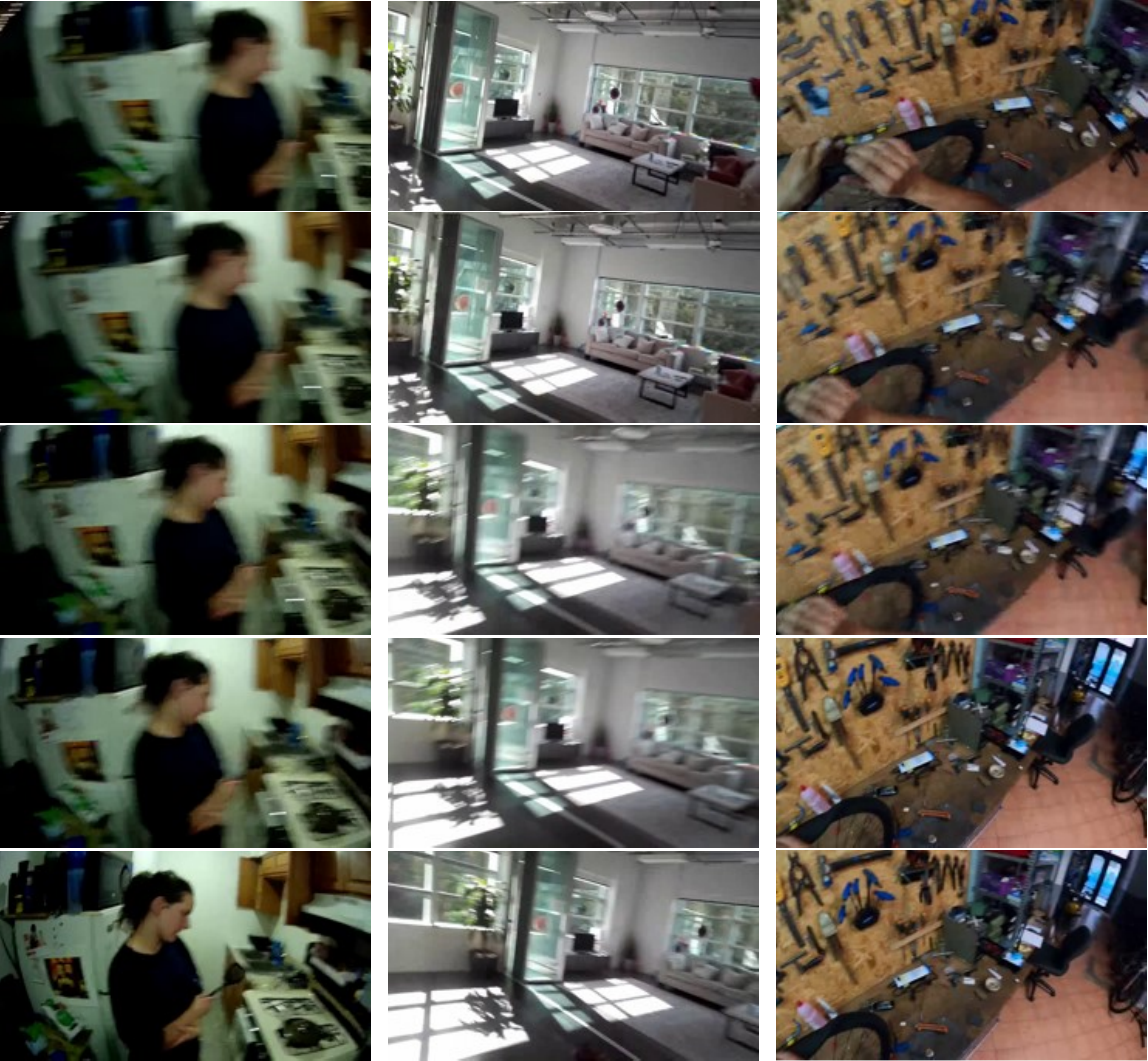}
		\caption{LR egocentric video frames.}
        \label{fig:head-1}
	\end{subfigure} 
        \hspace{0.02\textwidth}
	\begin{subfigure}[c]{0.6\textwidth}
		\centering
		\includegraphics[width=\textwidth]{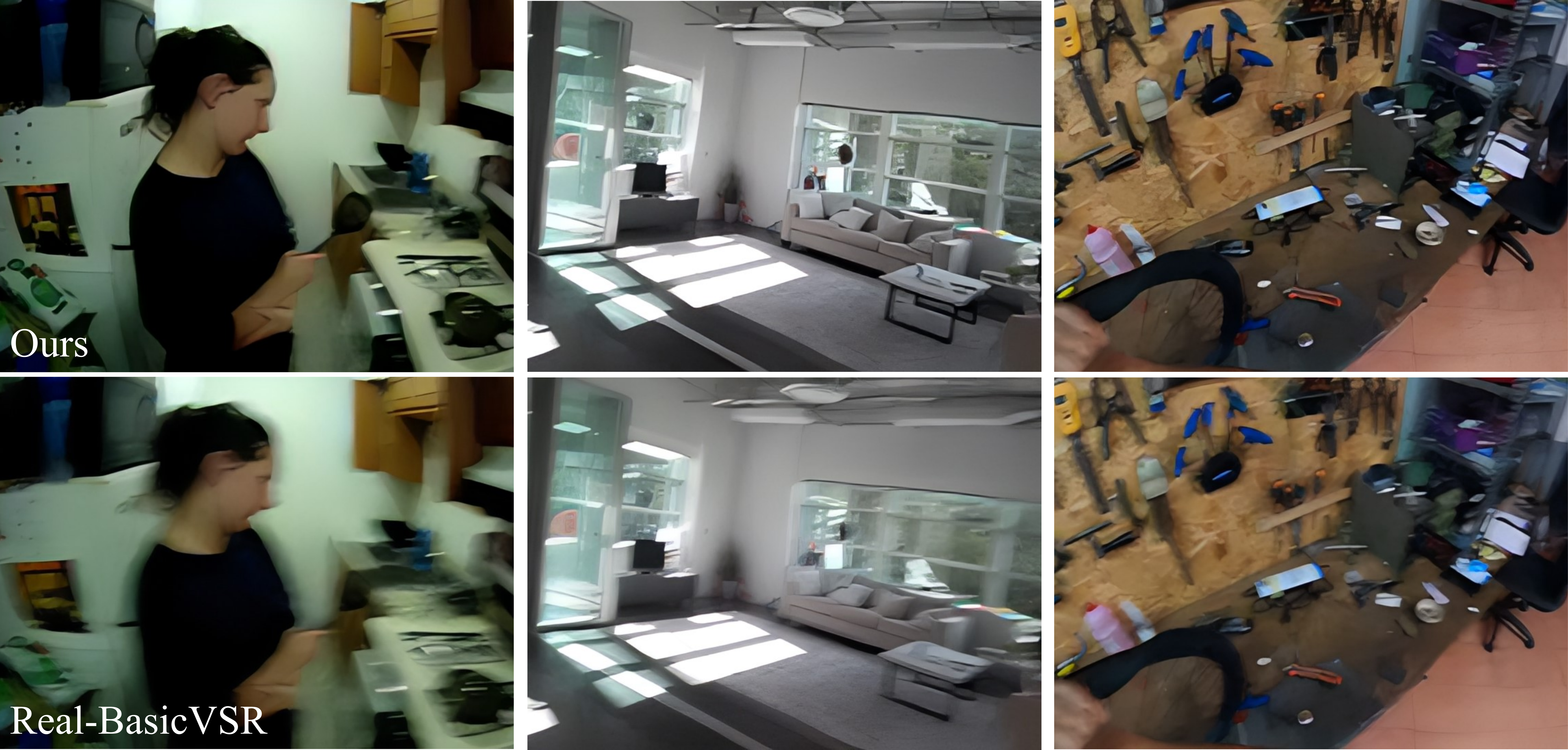}
		\caption{VSR results  from our method (top) and Real-BasicVSR~\cite{chan2022realBasicVSR} (bottom).}
        \label{fig:head-2}
	\end{subfigure}
  
   \caption{Egocentric videos generally contain temporally changing visual scenes with view changes and geometric deformations due to ego-motion, resulting in severe motion blur (see (a)). Our EgoVSR can handle motion blur in low-resolution (LR) frames and reconstruct photo-realistic high-resolution (HR) egocentric video frames.
   }
   \label{fig:head}
\end{figure*}

\begin{abstract}

   Due to the limitations of capture devices and scenarios, egocentric videos frequently have low visual quality, mainly caused by high compression and severe motion blur. With the increasing application of egocentric videos, there is an urgent need to enhance the quality of these videos through super-resolution. However, existing Video Super-Resolution (VSR) works, focusing on third-person view videos, are actually unsuitable for handling blurring artifacts caused by rapid ego-motion and object motion in egocentric videos. To this end, we propose EgoVSR, a VSR framework specifically designed for egocentric videos. We explicitly tackle motion blurs in egocentric videos using a Dual Branch Deblur Network (DB$^2$Net) in the VSR framework. Meanwhile, a blurring mask is introduced to guide the DB$^2$Net learning, and can be used to localize blurred areas in video frames. We also design a MaskNet to predict the mask, as well as a mask loss to optimize the mask estimation. Additionally, an online motion blur synthesis model for common VSR training data is proposed to simulate motion blurs as in egocentric videos. In order to validate the effectiveness of our proposed method, we introduce an EgoVSR dataset containing a large amount of fast-motion egocentric video sequences. Extensive experiments demonstrate that our EgoVSR model can efficiently super-resolve low-quality egocentric videos and outperform strong comparison baselines. Our code, pre-trained models and data can be found at https://github.com/chiyich/EGOVSR/.
   
\end{abstract}

\begin{IEEEkeywords}
   Super-Resolution, Egocentric Videos, Video Deblurring, Motion-blur Synthesis.
\end{IEEEkeywords}

\section{Introduction}

Egocentric videos are captured from a first-person perspective, usually by a wearable camera such as a GoPro or smart glasses. These cameras are often smaller and lighter than traditional cameras, which makes them more convenient for capturing everyday experiences. However, they tend to have lower-resolution sensors, which limit their image quality. Additionally, hardware upgrades to increase the resolution can be expensive, and recording high-resolution (HR) videos often result in shorter battery life~\cite{Gopro}.

Video super-resolution (VSR) techniques~\cite{kappeler2016VSRCNN,caballero2017VESPCN,wang2019edvr,chan2022realBasicVSR} can enhance the quality of videos by increasing their resolution. To address the limitations, one straightforward idea is to apply VSR to egocentric videos to enhance the visibility of visual details and increase the level of immersion for the viewer.  However, egocentric VSR is more challenging. Due to the wearer's head movements, there are frequent perspective changes and a large amount of ego-motion in egocentric videos, which causes severe motion blurs (see Fig.~\ref{fig:head-1}). In addition, egocentric video is often accompanied by compression and noise due to the highly dynamic and challenging environment with many moving objects and lighting changes. The current state-of-the-art VSR approaches are not sufficient to address these challenges.


The existing VSR approaches~\cite{kappeler2016VSRCNN,caballero2017VESPCN,tian2020tdan,jo2018DUF,wang2019edvr} typically assume that input LR video frames are clean without any noises and motion blurs. Obviously, they cannot be directly applied to handle real low-quality egocentric videos. Very recently, Chan \etal~\cite{chan2022realBasicVSR} began to investigate real-world VSR, in which  a higher-order degradation model is introduced for super-resolving real scenes. However, in addition to common degradation (\eg image blur, noise, and compression), motion blurs are ubiquitous in egocentric videos, and the real-world VSR approach for third-person videos cannot handle the motion blurs, resulting in unsatisfactory results (see Fig.~\ref{fig:head-2}).

To tackle the technical challenges, in this paper, we propose EgoVSR, a novel method for high-quality egocentric VSR that can simultaneously handle motion blur and other visual degradations, generating clean HR egocentric video frames from input low-quality frames. To the best of our knowledge, this is the first study to focus on a video restoration problem for egocentric videos. Upon the EgoVSR framework, we propose a Dual Branch Deblur Network (DB$^2$Net) to explicitly address motion blurs in input video frames, and features from the network are used to reconstruct HR egocentric video frames. To guide the
DB$^2$Net learning, we introduce a blurring mask that can localize blurred areas in video frames. We also design a MaskNet to predict the mask, as well as a mask loss to
optimize the mask estimation. Since real-world egocentric videos contain a random mixture of multiple degradations, a dataset with paired clean and complex degraded video clips is needed for egocentric VSR training. Thus, we design an online motion blur synthesis model to simulate data with a mixture of motion blur and other degradations. 

To validate the effectiveness of our proposed model, we build an egocentric VSR evaluation dataset by sampling video sequences from Ego4D~\cite{Ego4D2022CVPR}, which contains diverse egocentric video clips with camera ego-motion, object motions, and motion blurs. Extensive experiments show that our EgoVSR can effectively reconstruct HR video frames and suppress strong motion blurs in real-world egocentric videos, outperforming strong comparison baselines. 



Our contributions are summarized as follows:
\begin{itemize}
   \item We propose EgoVSR for egocentric video super-resolution, in which a blurring mask extraction module and a Dual-Branch Deblur Network are introduced to explicitly handle motion blurs. To the best of our knowledge, this is the first work on egocentric VSR.
   \item We propose an online motion blur synthesis model that can be easily integrated into the existing VSR degradation pipeline without the constraints of capture devices and camera parameters. 
   \item We construct an EgoVSR dataset from Ego4D\cite{Ego4D2022CVPR} to serve as a benchmark for evaluating egocentric VSR models. Our extensive experiments on this dataset show that our proposed model well restores real-world egocentric videos and outperforms recent VSR approaches.
 \end{itemize}

\section{Related works}
\subsection{Egocentric Vision} 
The rise of augmented reality (AR) and metaverse has brought extensive attention to egocentric video research. Thanks to the development of large-scale egocentric video datasets such as EPIC-KITCHENS~\cite{Damen2018EPICKITCHENS}, EgoCom~\cite{northcutt2020egocom}, and Ego4D~\cite{Ego4D2022CVPR}, more and more researchers have been investigating egocentric videos. Various egocentric vision tasks have been explored, including activity recognition~\cite{kazakos2019epic,li2021ego}, human-object interaction~\cite{cai2016understanding,damen2016you}, and video summarization~\cite{del2016summarization,lee2012discovering}. However, existing works have mainly focused on solving high-level tasks, while low-level tasks aimed at improving video quality for egocentric videos have not yet been explored. This work represents the first attempt to study the egocentric VSR task.


\subsection{Video Super-Resolution} 
VSR aims to restore high-quality HR frames by multiple LR frames. 
Most of the existing VSR approaches
\cite{kappeler2016VSRCNN,caballero2017VESPCN,jo2018DUF,xue2019ToFlow,wang2019edvr,tian2020tdan,
isobe2020RSDN,isobe2020video,isobe2020revisiting,chan2021basicvsr,
cao2021videoTrans,chan2021understanding,chan2022basicvsr++,wen2022video,liu2022TTVSR,isobe2022ETDM}
are based on BI (Bicubic+Down) downsampling or BD (Blur+Down) downsampling
and rarely consider restoration of real degradation, which leads to unsatisfactory restoration results in the egocentric video.
RLBSR~\cite{faramarzi2016blind} using blind deconvolution method to super-resolve real-life videos.
RealVSR~\cite{yang2021realVSR} introduced a real-world mobile-taken dataset,
DBVSR~\cite{pan2021DBVSR} used blind kernel estimation to restore degraded videos,
Real-BasicVSR~\cite{chan2022realBasicVSR} introduced high-order degradation models to improve the restoration in real scenes, but 
they still struggle with motion blurs in egocentric videos. In contrast, our EgoVSR model simultaneously handles motion blurs and other degradations in real-world egocentric videos, making it a promising solution for first-person videos.


\subsection{Real-world Super-Resolution} 
Considerations on degradation models have been extensively studied in image super-resolution.
Many works\cite{bell2019kernelGAN,gu2019IKC,luo2022dcls,xia2023meta,sun2023hybrid} have investigated the super-resolution based on limited degradation kernels.
RealSR~\cite{ji2020realSR} considered learning degradation kernels from real scene images and applying them in the generation of image pairs.
BSRGAN~\cite{zhang2021BSRGAN} first introduced the concept of degradation model to train the network by superposition of multiple degradations.
Real-ESRGAN~\cite{wang2021realESRGAN} introduced a higher-order degradation model to further improve the capability of simulating natural degradation.
BMDSRNet~\cite{niu2021blind} investigated the single image deblurring and super-resolution by learning dynamic spatio-temporal information.
However, the limited information in a single image makes these methods incapable of handling real degradation in egocentric videos.

\subsection{Deblurring}
Deblurring can remove blur (\eg defocus, gaussian, and motion blur) from images and videos and produces clear results.
Many single-image deblur methods use one blurred input to restore clear image.
DeFusionNET~\cite{tang2019defusionnet} and DID-ANet~\cite{ma2021defocus} estimate the defocus blurring map with additional defocus map supervision, but they can not be used to estimate motion blur.
BANet~\cite{tsai2022banet} introduces an unsupervised blur-aware attention to capture motion area, and XYDeblur~\cite{ji2022xydeblur} uses two decoders to estimate two orthogonal residual blur maps. Those unsupervised methods need independent training and are incapable of complex egocentric videos.
For video deblurring, most of the works\cite{nah2017GoPro,lin2022FGST,su2017DVD,li2021arvo,zhou2019STFAN,pan2020TSP,zhang2022STDAN} directly capture multi-frame information without estimate blur mask, and they studied offline synthesized datasets that use high frame rate cameras to synthesize low frame rate blurred videos and corresponding ground truths (\eg GoPro~\cite{nah2017GoPro}, DVD~\cite{su2017DVD} and REDS-Blur~\cite{nah2019REDS} dataset).
Apart from this, there are also some works studying the real blur-clear image pairs. RealBlur~\cite{RealBlur} used the beam splitter to acquire multi-camera images, and RSBlur~\cite{RSBlur} analyzed the multi-camera data and then proposes a new synthesis model related to the camera ISP.
However, these offline-synthesized datasets lead to difficulties in controlling the degree of different degradations in super-resolution tasks, which limit the diversity of training data.
Moreover, many methods acquired data related to camera parameters and devices, which also limits their synthesis model when applied to different datasets. Different from them, we propose a flexible online motion blur synthesis model.

\section{Method}
\begin{figure*}[htb]
   \begin{center}
   \includegraphics[width=\linewidth]{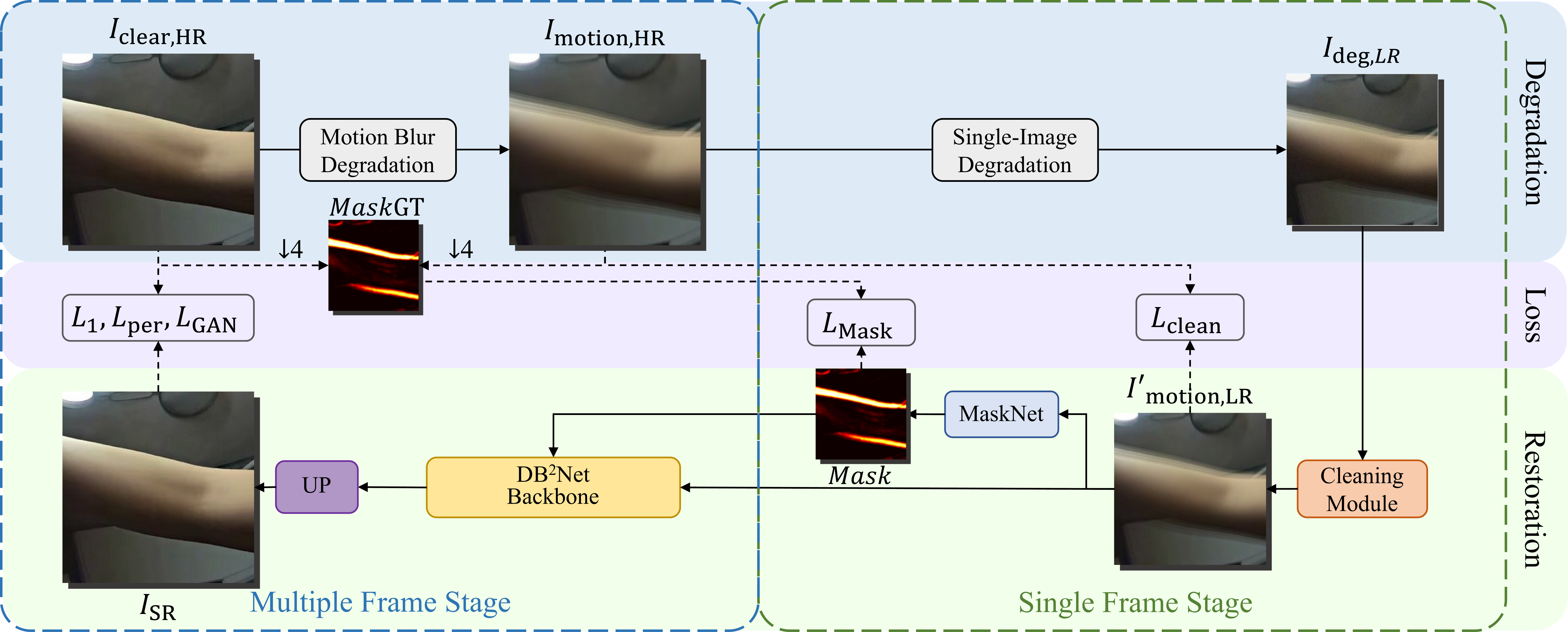}
   \end{center}
    
   \caption{Overall workflow of our EgoVSR. In the framework, we synthesize motion blur and other single-image degradations to generate paired training data and adopt a DB$^2$Net as the backbone model to perform blurring-aware context propagation for super-resolving LR frames, in which a MaskNet is learned to capture motion-blurred visual regions.  
   The trained EgoVSR model can be generalized to handle real-world egocentric videos.}
   \label{fig:overall}
\end{figure*}
\subsection{Overview}

The proposed EgoVSR establishes a high-quality egocentric video super-resolution model. 
The overall workflow of our method is shown in Fig.~\ref{fig:overall}.
The entire EgoVSR framework can be divided into two stages: a single-frame restoration stage and a multi-frame restoration stage. In the first stage, the single-frame image restoration stage is used to handle image degradation, such as compression, noise, and Gaussian blur. Then, the maskNet is adopted to extract blurring masks for each video frame. In the second stage, the cleaned LR frames and their corresponding blurring masks are fed into the DB$^2$Net backbone. The backbone learns the de-motion blur process through dual-branch deblurring, multi-frame alignment, propagation, and upsampling to generate HR results.

\subsection{Degradation Model}

Our analysis of degradation in the egocentric video revealed that it is essential to synthesize a random mixture of degradation in VSR training data. Importantly, the degradation process should include motion blur. To incorporate motion blur synthesis into the existing VSR degradation pipeline, we propose an online multi-frame motion blur synthesis model with several random parameters. The model can use motion information between multiple frames to synthesize frames with various degrees of motion blur, and can be applied to common VSR datasets without the limitations of capture devices and parameters.
The model has three random parameters, including the length of synthesized frames $N$, the stacking coefficient $r$, and the synthesis probability $p$. 
Here, $N$ denotes the total number of neighboring frames involved in the synthesis, $r$ serves as the coefficient of linear stacking among nearby frames, and $p$ controls the percentage of frames with motion blur. We use the following equation to synthesize the blurred frame $I^{(t)}_{motion,HR}$ corresponding to each frame $I^{(t)}_{clear,HR}$ according to the probability $p$:
\begin{equation}
\begin{aligned}
I^{(t)}_{motion}=\sum_{j=t-\lfloor N/2 \rfloor}^{t+\lfloor N/2 \rfloor}{r \cdot I^{(j)}_{clear}}+(1-N \cdot r)\cdot I^{(t)}_{clear}
\end{aligned}
\end{equation}
By applying a linear stack to nearby frames, motion areas are blurred while clear areas are preserved. The stacking coefficient $r$ and the length of the frames $N$ control the degree of blur, thus increasing the diversity of the synthesized data.
For any video frame clip, the algorithm summarized in Alg.~\ref{alg:alg1} can be used to synthesize \textbf{first-order} motion blur.
\begin{algorithm}[H]
   \caption{Online Motion Blur Synthesis Model}\label{alg:alg1}
   \begin{algorithmic}
   \STATE \textbf{Input:} clear video frames $\{I^{(1)}_{clear},...,I^{(n)}_{clear}\}$
   \STATE \textbf{Output:} blurred video frames $\{I^{(1)}_{motion},...,I^{(n)}_{motion}\}$
   \STATE 
   \STATE 1, Randomly choose the synthesis parameters: 
   \STATE \hspace{1cm} $ N\in \{3,5,7\} $ , $ r\in[0,1/N] $ , $ p\in(0.5,1] $
   \STATE 2. For the intermediate video frames $ \{ I^{(i)}_{clear},i \in \left[ \lceil N/2 \rceil,n- \lceil N/2\rceil \right]  \}$, synthesis according to the following way by the probability $p$: 
   \STATE \hspace{1cm} $I^{(i)}_{motion}=\sum_{j=i-\lfloor N/2 \rfloor}^{i+\lfloor N/2 \rfloor}{r \cdot I^{(j)}_{clear}}+(1-N \cdot r)\cdot I^{(i)}_{clear} $ 
   \STATE 3. For all synthesized video frames $I^{(i)}_{motion}$, bound its range to [0,1]:\\
   \STATE \hspace{1cm} $I^{(i)}_{motion}=clamp(I^{(i)}_{motion},0,1)$ \\
   \end{algorithmic}
\end{algorithm}
Additionally, we can get the \textbf{second-order} motion blur data by applying this model to the pre-synthesized deblur dataset (\eg REDS-Blur~\cite{nah2019REDS}).
Such data includes the real blurring scenarios in the pre-synthesized dataset, and provides severe blurring that is more applicable to egocentric data.

 After synthesizing motion blur frame $I^{(t)}_{motion,HR}$ for each timestamp $t$, we introduce the higher-order degradation model used in Real-ESRGAN~\cite{wang2021realESRGAN} 
 and Real-BasicVSR~\cite{chan2022realBasicVSR} to simulate more realistic degraded frame $I^{(t)}_{deg,LR}$, which helps to restore higher-quality results.

\subsection{Network Architecture}
\begin{figure*}[htb]
   \begin{center}
   \includegraphics[width=\linewidth]{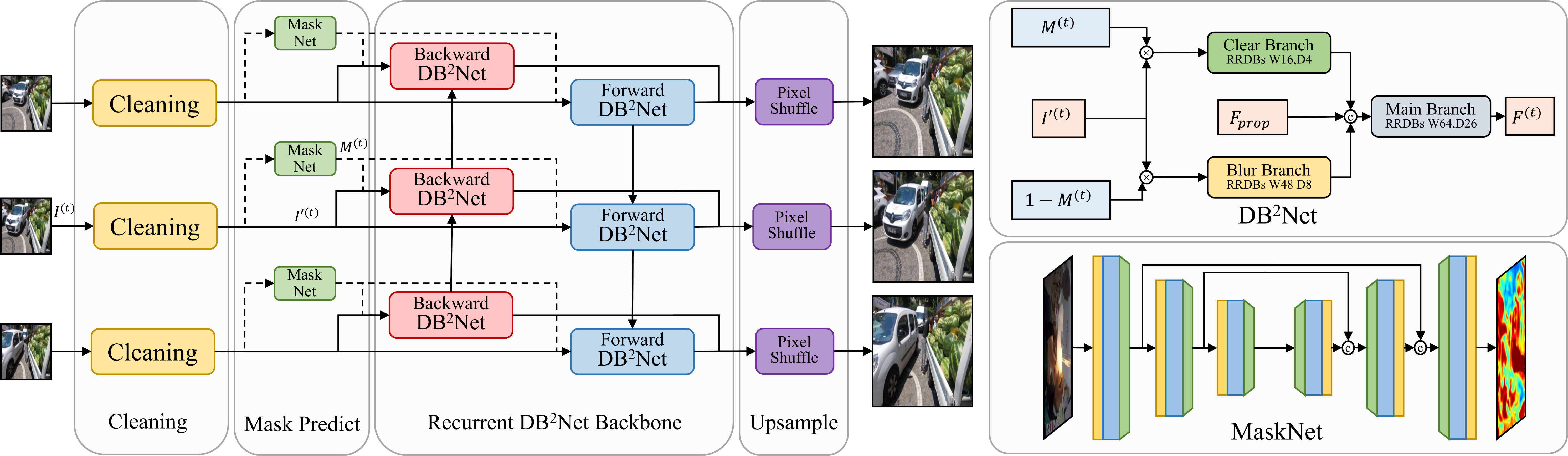}
   \end{center}
    
      \caption{Network architecture of the proposed EgoVSR framework.
      The refined and cleaned images are passed through MaskNet to extract the blurring mask, then the cleaned image and mask are fed into the DB$^2$Net backbone for bidirectional propagation and finally obtain the upsampled results.
      Note that our network is trained end-to-end.}
   \label{fig:arch}
    
\end{figure*}

The architecture of our EgoVSR network is illustrated in Fig.~\ref{fig:arch}, which is composed of four parts: the Cleaning Module, Blurring Mask Extraction Module, DB$^2$Net Backbone, and Upsampling Module.
Firstly, the same Cleaning Module as used in Real-BasicVSR~\cite{chan2022realBasicVSR} is applied to pre-clean the single-image degradation in the video frames $I_{deg,LR}$.
Then, the Blurring Mask Extraction Module is utilized to identify the blurred areas in the frames.
Next, the cleaned video frames $I'_{motion,LR}$ and their corresponding blurring masks $M$ are fed into the DB$^2$Net Backbone to extract the final features.
These features are then passed through the PixelShuffle~\cite{shi2016PixelShuffle} upsampling layer to generate the final VSR results $I_{SR}$.

\subsection{Blurring Mask Extraction}\label{sec:maskloss}
The blurring mask extraction module proposed in our method consists of two main components: the mask extraction network (MaskNet) and the mask loss function.
The MaskNet is responsible for estimating the blurred area in the low-resolution (LR) frame and generating a blurring mask that will guide the subsequent network in the deblurring process.
Meanwhile, the mask loss function is designed to estimate the ground-truth mask from the synthesized blurred data and clear data, which then supervises the learning of the network.
Together, these two components work in tandem to ensure that the blurring mask is accurately estimated and applied to the deblurring process, ultimately improving the quality of the final VSR results.

The U-Net-like MaskNet, illustrated in Fig.~\ref{fig:arch}, consists of three downsampling layers and three upsampling layers. Each sampling layer contains a front convolution layer, an up/down sampling operation, and four layers of residual CNN structures.
When processing each input frame $I'_{motion,LR}$~(Here, we ignore the timestamp $(t)$ for a clearer expression), it is first passed through the downsampling layer to obtain the downsampling features $M_{i,down}$ for $i=1,2,3$.
Subsequently, each upsampling layer concatenates the previous output $M_{i,up}$ (in reverse order for $i=3,2,1$) with the corresponding skip-connection features $M_{i,down}$, and performs a deconvolution operation with stride 2 to obtain the output of the current layer.
Finally, the output of the last layer is normalized by the softmax function to obtain a single-channel mask $M$ with the same size as the input image.

To improve the interpretability of the mask $M$ and scale its distribution within the range of [0,1], we introduce a mask loss function. Specifically, we define the mask values of a clear area of the image as 1 and the mask values of the fully motion-blurred area as 0. The blurring mask of a given frame is generated using the following method:

We first obtain the residual map $M_{res} \in R^{1\times H_{LR}\times W_{LR}}$ by computing the channel-wise Mean Square Error~(MSE) between the downsampled clear frame $I_{clear,HR\downarrow 4}$ and synthetic motion blur frame $I_{motion,HR\downarrow 4}$, and multiplying it by the magnification factor $k$:
\begin{equation}
M_{res}=k\cdot MSE(I_{motion,HR\downarrow 4},I_{clear,HR\downarrow 4} )
\end{equation}
The blurred areas present in the image are marked by the residuals between the blurred image and the clear image. However, as the residuals may be small, we multiply by $k$ to amplify the difference in detail.
Next, to limit its range to [0, 1] and make the mask more interpretable, we clip it with the function $clamp(\cdot)$ to obtain the clamped residual distribution map $ M_{clamp}=clamp(M_{res},min=0,max=1)$.
To further soften the edges in the mask and remove noise, we apply a Gaussian blur kernel with kernel size=7 and $\sigma$=3 to the $M_{clamp}$ to obtain the final mask ground-truth $M_{GT}$.
In this way, we obtain the ground-truth of the blurring mask from the residuals of the clear and blurred images by differencing, filtering, and softening.

Since the motion blur distribution in the data is non-uniform, we only use mask loss for frames with more blur (a higher percentage of zeros in the mask). Specifically, we compute the mean mask value for each frame and only compute $L_{mask}$ for the frames with a mean mask value under a threshold of 0.6.
\begin{gather}
   M_{th}= \left(mean(M_{GT} )<0.6\right), \\
   L_{mask}=||\left(M_{GT}-MaskNet(I_{LR})\right)\cdot M_{th}||_1.
\end{gather}

\subsection{Dual-Branch Deblur Network~(DB$^2$Net)}
There are both clear areas and motion-blurred areas in each video frame.
Balancing the restoration results between clear and motion-blurred areas can be a challenge when using the same parameters for inference. This can result in artifacts in clear areas or inadequate deblurring in blurred areas. To address this, we propose a Dual-Branch Deblur network (DB$^2$Net), which processes both the clear and blurred areas of the image in two separate branches and then performs feature fusion to yield clear frame features. This allows our network to propagate trustable clear features in feature propagation. 
The architecture of this network is shown in Fig. \ref{fig:arch} (see top right).
Akin to previous VSR methods~\cite{chan2021basicvsr}, we stack multiple DB$^2$Net for bidirectional~(forward and backward) propagation.

Firstly, the cleaned LR images $I'^{(t)}_{motion,LR}$ is element-wise multiplied with the estimated mask $M^{(t)}$ and $1-M^{(t)}$ to obtain two masked inputs, one for the clear areas and one for the blurred areas in the frame. These masked inputs are then processed by two different branches separately.
Among them, the clear and blur branches are a Residual-in-Residual Dense Blocks~(RRDBs)~\cite{wang2018esrgan} architecture:
\begin{gather}
   F_{clear}^{(t)}=ClearBranch(I'^{(t)}_{motion,LR}\cdot M^{(t)}) \\
   F_{blur}^{(t)}=BlurBranch\left(I'^{(t)}_{motion,LR}\cdot (1-M^{(t)}) \right) 
\end{gather}
Then, we concatenate the dual branch results with warped propagation features $F_{prop}$ and pass them into the main branch to get the final clear features $F^{(t)}$:
\begin{gather}
   F^{(t)}=MainBranch\left(cat(F_{clear}^{(t)},F_{motion}^{(t)},F_{prop} )\right)
\end{gather}
For the forward and backward propagation features $F^{(t)}$, we denote them as $F_{forward}^{(t)}$ and $F_{backward}^{(t)}$. 
Here, $F_{prop}$ is the feature of the previous frame and the next frame in the forward and backward propagation, respectively. 
\begin{gather}
   F_{prop} = \left\{
      \begin{aligned}{}
      warp(F^{(t-1)}_{forward},flow(F^{(t-1)}\rightarrow F^{(t)}))\\
      warp(F^{(t+1)}_{backward},flow(F^{(t+1)}\rightarrow F^{(t)}))\\
      \end{aligned} \right. 
\end{gather}
$warp$ denotes feature warping operation, and $flow$ is calculated by pre-trained optical flow SpyNet~\cite{ranjan2017optical}.
Finally, we use a pixelshuffle layer to generate the super-resolution results:
\begin{gather}
   I_{SR}^{(t)}=Upsample\left(cat(F_{forward}^{(t)},F_{backward}^{(t)})\right)
\end{gather}

\subsection{Loss Function}

The objective of the optimization is to tackle both single-image degradation and motion blur while generating photo-realistic restoration results. To achieve this, we incorporate the cleaning loss $L_{clean}$ from Real-BasicVSR \cite{chan2022realBasicVSR} to better handle single-image degradation. We adjust the supervision of this loss on frames with motion blur to focus on restoring single-image degradation:
\begin{equation}
    L_{clean} = ||I_{motion,HR\downarrow4}-I'_{motion,LR}||_1
\end{equation}
Additionally, we introduce the mask loss $L_{mask}$ to supervise the training of MaskNet, as described in Sec.~\ref{sec:maskloss}.
Similar to GAN-based SR models~\cite{chan2022realBasicVSR,wang2021realESRGAN}, we use L1 loss, perceptual loss $L_{per}$~\cite{johnson2016perceptual}, and GAN loss $L_{GAN}$~\cite{goodfellow2020generative} to optimize the whole network. 
Our full optimization objective function is defined as:
\begin{gather}
   L_{total} = L_1 + L_{prep} + L_{clean}+ \lambda_{1} L_{GAN}+ \lambda_{2} L_{mask}   
\end{gather}
where ${\lambda_{1}}$ and $\lambda_{2}$ are hyperparameters to balance the loss items, and the weights not specifically noted are empirically set to 1. 
In our implementation, we empirically set them $5\times 10^{-2}$ and $2\times 10^{-1}$, respectively.

\section{Experiments}

\subsection{Experiments Setup}\label{sec:setup}
\noindent
\textbf{EgoVSR dataset.} To validate the performance of our method under egocentric videos, we sampled video clips from Ego4D~\cite{Ego4D2022CVPR} and built the EgoVSR dataset.
The dataset includes 500 video sequences with fast motion, and the train, test, and validation sets are divided by 90:6:4. 
To unify the resolution, we first uniformly scaled and cropped the video to 960$\times$544 for the HR clips. Then, we use bicubic downsampling with a factor of 4 to get LR videos.
After all videos are of the same resolution, we crop video clips from the videos according to the average optical flow movement of the frames.
In generating the training set, we randomly sample 10 video clips from these videos to obtain the clearest possible data. Then, we randomly sample 1 to 4 fast-motion clips from each video for test and validation sets. Each clip in the training set includes 15 frames, while the test and validation sets include 7 frames. 
The settings for each subset are shown in Table~\ref{tab:egovsr-setting}.
Note that we did not add any additional degradation into these LR egocentric video frames. They contain real captured degradation, including motion blur and others (\eg, compression artifacts).
\begin{table}[htbp]
  \centering
  \setlength\tabcolsep{4pt}
  \caption{Settings of EgoVSR dataset.}
    \begin{tabular}{c|ccc|c}
    \toprule
    Subset & Videos & Clips & Frames & Resolution \\
    \midrule
    Train & 4500   & 45000 & 315000 & \multirow{3}{*}{960$\times$544} \\   
    Valid  & 200 & 682   & 4774  &  \\   
    Test & 300  & 1042  & 7294  &  \\
    \bottomrule
    \end{tabular}%
  \label{tab:egovsr-setting}%
\end{table}%

\noindent
\textbf{Implementation Details.} 
In DB$^2$Net, The number of RRDBs for the clear branch is set to 4 with channels 16, for the blur branch is set to 8 with channels 48, and for the main branch is set to 26 with channels 64. 
In MaskNet, we set the channels of the U-Net architecture to [16,32,64], and each downsampling and upsampling layer is followed by four layers of residual convolution.

\noindent
\textbf{Training Settings.} 
As Real-BasicVSR~\cite{chan2022realBasicVSR}, we used the REDS~\cite{nah2019REDS} dataset for training and appended the training results of the EgoVSR dataset for comparison.
We adopt Adam optimizer~\cite{kingma2014adam} to train our model, with different learning rates in 2 stages using NVIDIA GeForce RTX 3090.
We use 64$\times$64 LR image patch for training, and each training segment contains 30 frames. 
We first train the PSNR-oriented model, including $L_1$, $L_{Mask}$ and $L_{clean}$, with a learning rate of $10^{-4}$ and batch size 4 for 200K iterations. 
Then, we further introduce $L_{GAN}$ and $L_{per}$ to train 250K iterations with batch size 2 and a learning rate of $5 \times 10^{-5}$ for the generator and $10^{-4}$ for the discriminator.

\noindent
\textbf{Evaluation Metrics.}
HR video frames are originally captured egocentric video frames in our EgoVSR dataset. As discussed, these HR frames usually contain compression artifacts and motion blurs, which cannot be directly used as groundtruth for evaluation. 
To address the issue, we adopt three non-reference IQA metrics, including NIQE~\cite{mittal2012NIQE}, PI~\cite{blau2018PI}, and NRQM~\cite{ma2017NRQM} to measure the quality of reconstructed HR egocentric video frames.

\subsection{Training Data Selection}\label{sec:ego-reds}
The EgoVSR training set consists of 45,000 clips randomly sampled from 4,500 videos, and each clip consists of 15 frames at 960$\times$544 resolution. We filter the data with slow motion to build the training set with the sharper images as possible.
Initially, we used the EgoVSR dataset for training, but the resulting model produced many artifacts and blurred edges in the restored frames. To address this issue, we adopted the REDS-VSR~\cite{nah2019REDS} training set, which has been used in existing VSR methods. Additionally, we synthesized motion blur and introduced single-image degradation into video frames to simulate real-world degradation commonly observed in egocentric videos.
The model trained on REDS resulted in excellent performance and significantly improved the quality of the restored HR frames in the EgoVSR test set over the model trained on EgoVSR. Furthermore, we experimented with two mixture methods, REDS+Ego~(REDS pre-training and EgoVSR fine-tuning) and Ego+REDS~(EgoVSR pre-training and REDS fine-tuning). However, the results do not show a significant improvement over the single REDS-trained model.
Experimental results are shown in Tab.~\ref{tab:ego-reds-exp}.
\begin{table}[h]
   \centering
   \setlength\tabcolsep{3pt}
   \caption{Test result under different training setting.}
      \scalebox{1}{\begin{tabular}{c|cccc}
     \toprule
     TrainSet     & REDS    & Ego   & REDS+Ego & Ego+REDS \\
   \midrule
     NIQE↓ & \textbf{4.9678} & 6.7135 & 5.1316 & 5.0301\\
     PI↓   & \textbf{4.0815} & 5.5600 & 4.2253 & 4.1890\\
     NRQM↑  & \textbf{6.8047} & 5.5935 & 6.6810 & 6.6522\\
     \bottomrule
     \end{tabular}}%
      
   \label{tab:ego-reds-exp}%
 \end{table}%
\begin{figure}[h]
   \begin{center}
   \includegraphics[width=\linewidth]{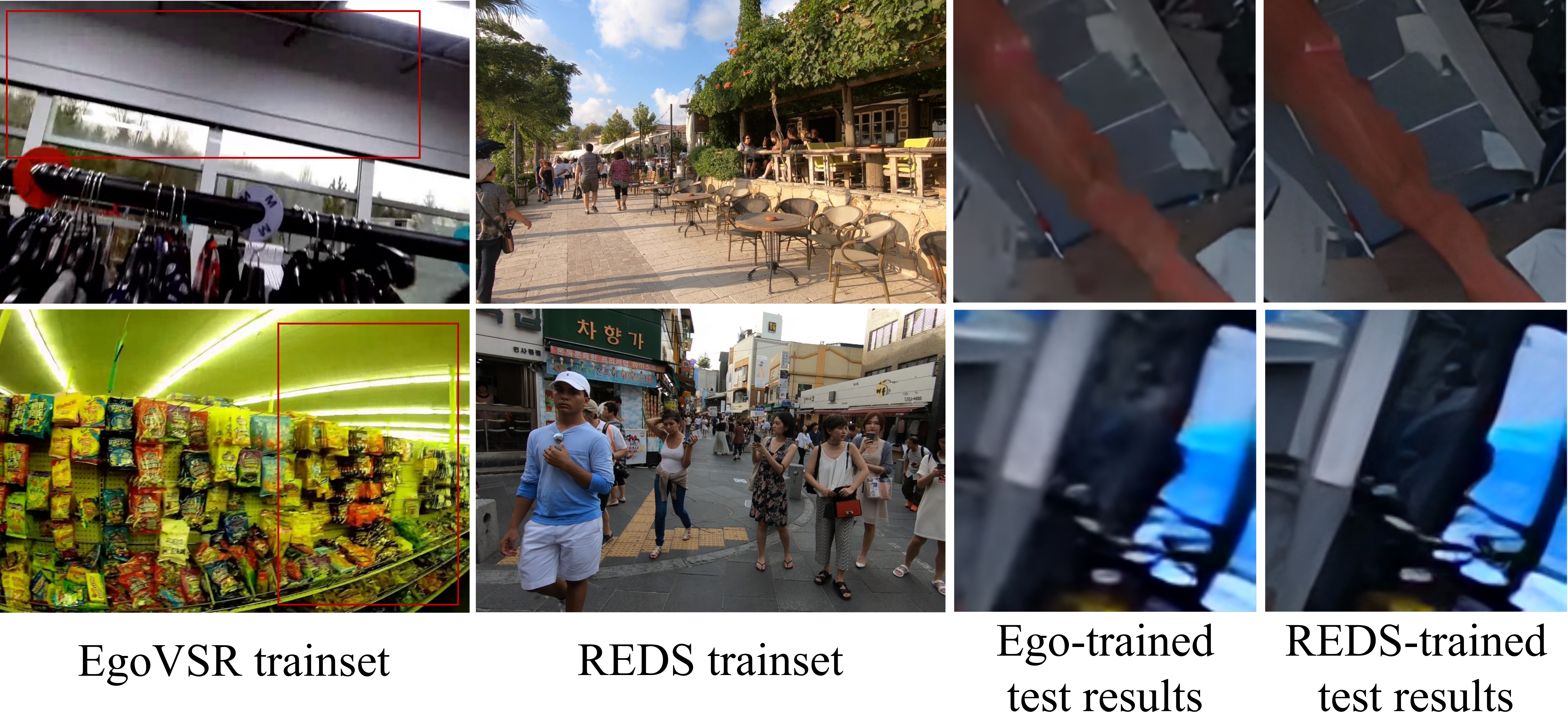}
   \end{center}
       
      \caption{Comparison of EgoVSR and REDS.}
   \label{fig:ego-reds-exp}
\end{figure}

To investigate why the performance of REDS transfer learning is higher than the native EgoVSR dataset, 
we show the frames in the two training sets and the cropped testing results under REDS training and EgoVSR training in Fig.~\ref{fig:ego-reds-exp}. 

It can be observed that the EgoVSR dataset suffers from high compression and motion blur, thus leading to the lack of clear and high-quality HR supervision.
As a result, the EgoVSR-trained model underperforms the REDS-trained one.
We finally selected the REDS dataset and added our degradation synthesis model (first-order and second-order) to address the egocentric VSR task. Besides, we also train our model on the original REDS-Blur datasets, please refer to Sec.~\ref{sec:abl}.

\begin{figure*}[tbp]
   \begin{center}
   \includegraphics[width=\linewidth]{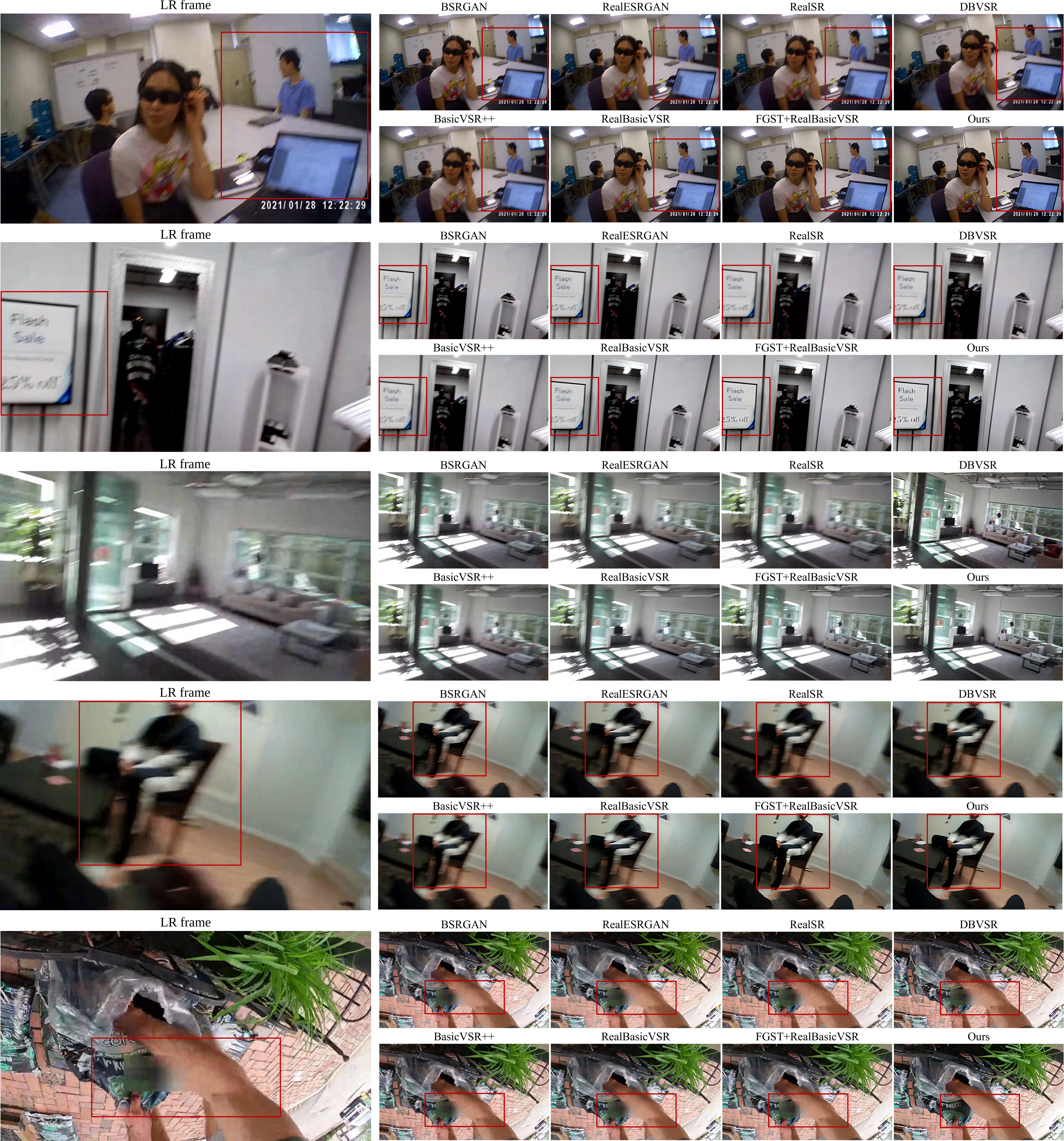}
     
   \end{center}
      \caption{Egocentric VSR results from different methods. Our approach can achieve excellent deblurring results and restore a clear image from the aggregation of nearby frames.}
   \label{fig:main-exp}
\end{figure*}

\begin{table*}[tbp]
   \centering
   \setlength\tabcolsep{3pt}
   \caption{Quantitative comparison of multiple methods. The best and the second-best results are \textbf{highlighted} and \underline{underlined}. 
   FLOPs are calculated with input size (7,3,128,128).
   }
   
      \scalebox{1}{\begin{tabular}{l|cccccccc}
\toprule
       
      & \makecell[c]{Real-ESRGAN\\\cite{wang2021realESRGAN}}
      & \makecell[c]{BSRGAN\\\cite{zhang2021BSRGAN}}
      & \makecell[c]{RealSR\\\cite{ji2020realSR}}
      & \makecell[c]{BasicVSR++\\\cite{chan2022basicvsr++}}
      & \makecell[c]{DBVSR\\\cite{pan2021DBVSR}}
      & \makecell[c]{Real-BasicVSR\\\cite{chan2022realBasicVSR}}
      & \makecell[c]{FGST\cite{lin2022FGST}+\\Real-BasicVSR} 
      & Ours \\
\midrule
     NIQE↓ & 6.1674 & 5.6144 & 5.6798 & 7.3134 & 7.3609 & 5.5375 & \underline{5.4791} & \textbf{4.9678} \\
     PI↓   & 5.0107 & 4.5483 & 5.1081 & 6.4523 & 6.5582 & 4.6283 & \underline{4.5213} & \textbf{4.0815} \\
     NRQM↑ & 6.1459 & \underline{6.5177} & 5.4635 & 4.4089 & 4.2445 & 6.2809 & 6.4365 & \textbf{6.8047} \\
\midrule
     FLOPs  & 2055.99G & 2055.99G & 2055.99G & 793.182G & 320.507G & 1077.71G & 1396.19G & 1116.33G \\
     
     Params  & 16.698M & 16.698M & 16.698M & 7.028M & 20.538M & 4.855M & 14.536M & 6.402M  \\
  \bottomrule
     \end{tabular}%
     }
      
   \label{tab:main-exp}%
 \end{table*}%

\subsection{Comparison to State of the Arts}\label{sec:main}
\noindent\textbf{Comparison Setting.} We conducted extensive experiments on the EgoVSR dataset (testset) to verify the effectiveness of our approach.
Six SR methods are used for comparison, 
including three single image SR methods Real-ESRGAN~\cite{wang2021realESRGAN},BSRGAN~\cite{zhang2021BSRGAN}, and RealSR~\cite{ji2020realSR}, 
and three VSR methods DBVSR~\cite{pan2021DBVSR}, BasicVSR++~\cite{chan2022basicvsr++}, and Real-BasicVSR~\cite{chan2022realBasicVSR}.
Besides these, we cascaded a SOTA video deblurring method FGST~\cite{lin2022FGST} with Real-BasicVSR for comparison. 
All the above methods use the officially provided pre-trained models. We use the official REDS-VSR-trained model for both DBVSR, BasicVSR++, and Real-BasicVSR, as well as our method in the comparison section.
Note that our second-order model introduce additional REDS-Blur datasets which may cause unfair compairson, so we selected the first-order model in this section, and we will discuss the second-order model later.

\noindent\textbf{Quantitative Results.} Results are shown in Tab.~\ref{tab:main-exp}.  Our EgoVSR outperforms all other methods in terms of the three metrics. These results suggest that real-world SR approaches can significantly enhance the performance of egocentric VSR, and adding video deblurring can provide an additional boost. Our unified egocentric VSR approach achieved the best overall performance compared to other methods, which validates the effectiveness of the proposed EgoVSR framework.

\noindent\textbf{Qualitative Results.} To further demonstrate the superiority of our method, we provide a qualitative comparison in Fig.~\ref{fig:main-exp}.
Through the aggregation of multi-frame information by DB$^2$Net and the use of blurring masks, EgoVSR is able to achieve excellent deblurring results while super-resolving.
Interestingly, the last frame shows the capability of our method in handling the mosaic area by aggregating temporal information, demonstrating the great potential of EgoVSR.

 \begin{figure}[h]
	\centering
	\includegraphics[width=\linewidth]{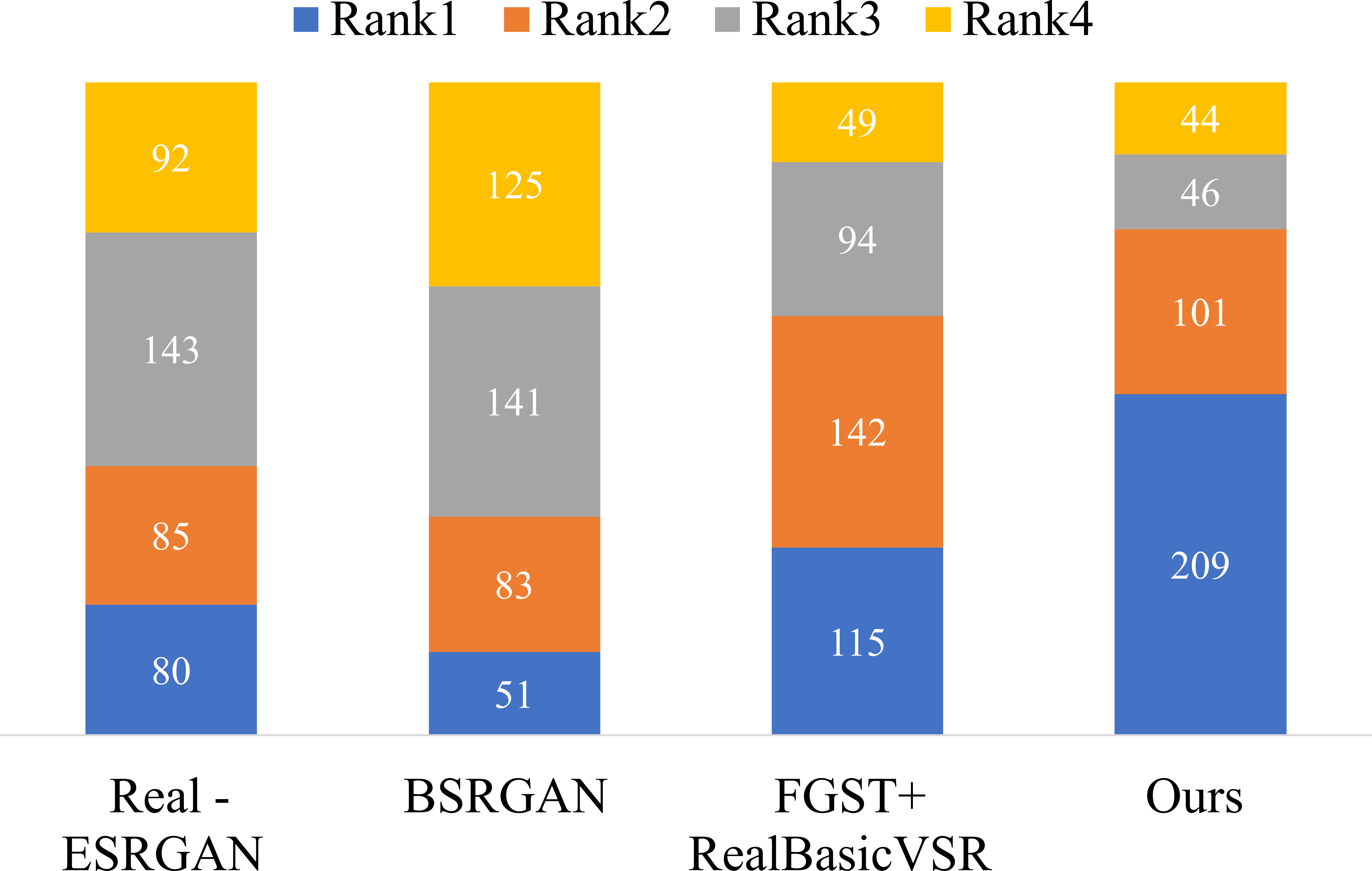}
   \caption{User study among 4 methods. Our method achieves the highest first-place rate (FPR) and the lowest average rank (AR).}
   \label{fig:UserStudy}
\end{figure}

\noindent\textbf{User Study.} Furthermore, we randomly selected 20 images from the EgoVSR dataset to conduct a user study.
We choose four methods (Real-ESGRAN~\cite{wang2021realESRGAN}, BSRGAN~\cite{zhang2021BSRGAN}, FGST~\cite{lin2022FGST}+Real-BasicVSR~\cite{chan2022realBasicVSR}, EgoVSR) and shuffled the results of different methods to ensure that users do not experience fatigue.
A total of 20 users blindly ranked the 20 sets of results of four methods as they preferred. 
Results of user study are shown in Fig.~\ref{fig:UserStudy}.
The average rank (AR↓) of the four methods were (2.62, 2.85, 2.19,1.81) among 1600 results, 
and the first-place rate (FPR↑) are (20.00\%, 12.75\%, 28.75\%, 52.25\%) among the 400 samples. The ratio sum is greater than 1 since the rankings can be tied.
It can be seen that our method also achieved the best performance in the user study.

\subsection{Ablation Study}\label{sec:abl}
\begin{table*}[t]
   \centering
   \caption{Qualitative ablation study of our method.}
      \scalebox{1}{\begin{tabular}{l|cccccc}
     \toprule
           & baseline~\cite{chan2022realBasicVSR} & model 1  & model 2  & model 3 & model 4 & model 5\\
     \midrule
     Motion Blur & \ding{55}     & Fisrt-Order     & Fisrt-Order  & REDS-Blur & Fisrt-Order  & Second-Order\\

     MaskNet+DB$^2$Net & \ding{55}  & \ding{55}   & \ding{51}  & \ding{51} & \ding{51} & \ding{51}\\

     MaskLoss & \ding{55}     & \ding{55}     & \ding{55}     & \ding{51} & \ding{51} & \ding{51} \\
\midrule
     NIQE↓   & 5.5375 & 5.4158 & 5.1241 & 5.0988 & 4.9678  &\textbf{4.6497} \\

     PI↓     & 4.6283 & 4.5214 & 4.4520 & 4.2780 & 4.0815  &\textbf{4.0480} \\

     NRQM↑  & 6.2809 & 6.3730 & 6.6436  & 6.5428 & \textbf{6.8047} & 6.5537 \\
     \bottomrule
     \end{tabular}%
     }
      
   \label{tab:ablation-exp}%
 \end{table*}%

In this section, we show the effects of each component in our EgoVSR, including the motion blur synthesis model, DB$^2$Net with MaskNet, and mask loss function. 
Since the input of DB$^2$Net depends on the result of MaskNet, the two cannot be used separately.
We choose Real-BasicVSR as our baseline model and sequentially apply the components above, then re-train each model as in Sec.~\ref{sec:setup}, named model 1 to model 5. 
To ensure a fair comparison, we reduce the number of channels and total RRDBs in DB$^2$Net compared to Real-BasicVSR backbone.
Table~\ref{tab:ablation-exp} shows the qualitative results of each model. 
By adding the motion blur synthesis, model 1 can better deal with motion blur.
When replacing the backbone network to DB$^2$Net, the restoration capability of the network (model 2) significantly increases.
When Mask Loss is added, the performance is further improved.

\begin{figure}[h]
   \begin{center}
   \includegraphics[width=\linewidth]{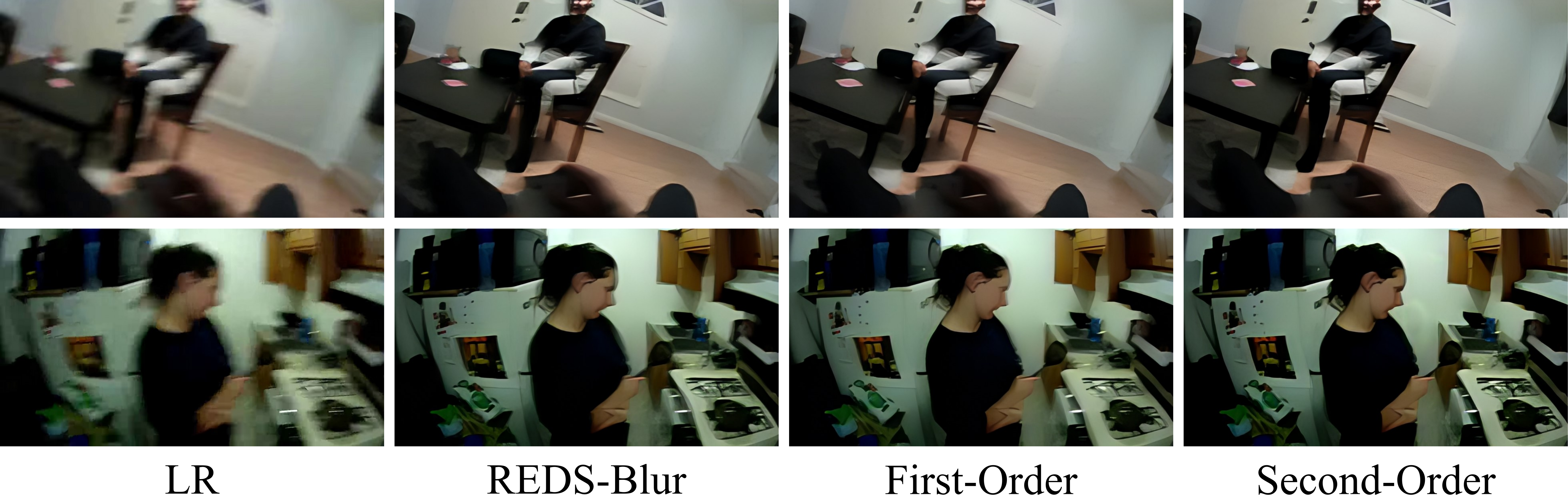}
   \end{center}
      \caption{Qualitative compairson between REDS-Blur trained model and our final model. Our model is able to perform better deblurring and obtain sharper super-resolution results.}
   \label{fig:reds-deblur}
\end{figure}

\begin{figure*}[h]
   \begin{center}
   \includegraphics[width=\linewidth]{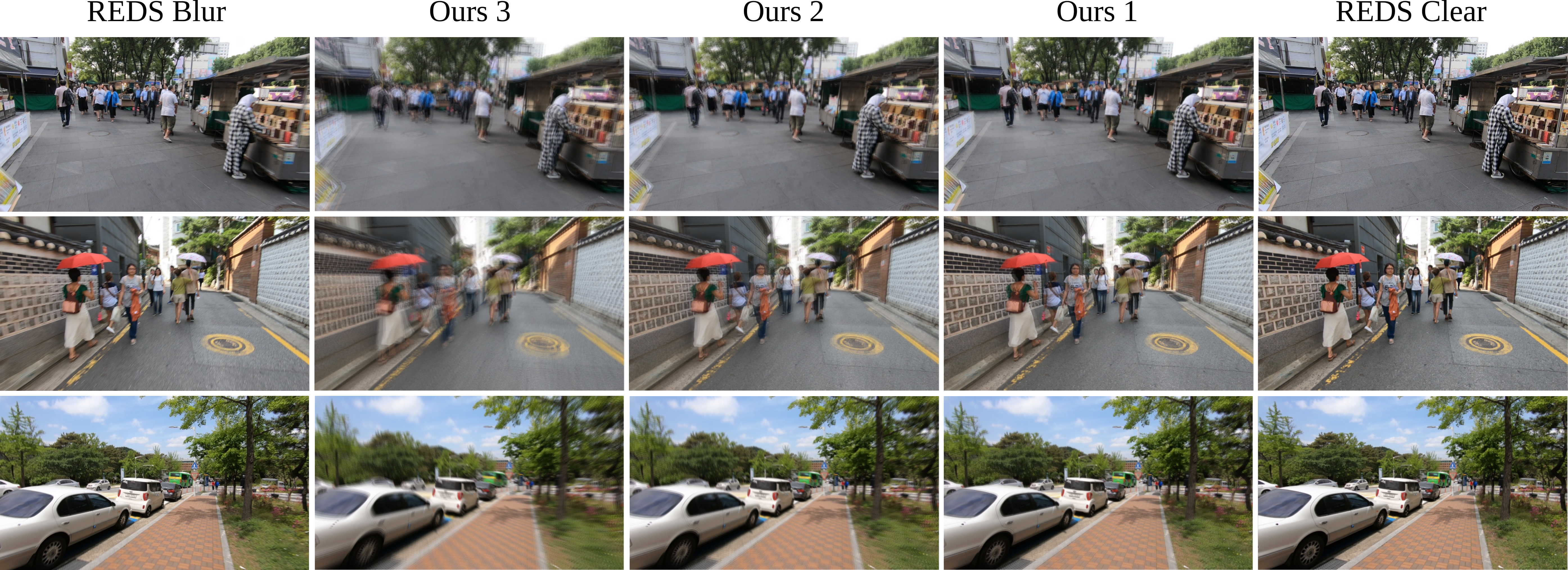}
   \end{center}
    
      \caption{Comparison of REDS pre-synthesized blurred images and our online synthesized images. Our model is capable of synthesizing different degrees of motion blur~(Ours 1-3), while the REDS pre-synthesized data only has a slight blur.}
   \label{fig:reds-blur-demo}
\end{figure*}

In addition, we can observe that the model trained on REDS-Blur dataset underperforms the model trained on our motion blur synthesis model, both first-order model and second-order model, as shown in Fig.~\ref{fig:reds-deblur}. And our model performs better deblurring and obtains sharper super-resolution results while applying second-order model. To demonstrate why the model trained on REDS-Blur dataset is not capable of restoring egocentric video well, we further show the visualization of REDS pre-synthesized blurred images and our online synthesized images in Fig.~\ref{fig:reds-blur-demo}.

It can be seen that the REDS-Blur dataset only has a slight blur, while our synthesis model can simulate different degree of motion blur~(Ours 1 to Ours 3 have increasingly motion blur). Since the motion blur in egocentric videos are more severe and diverse than third-view videos, the original REDS pre-synthesized images are not suitable for training the EgoVSR. Since we applied our motion blur synthesis model to the REDS dataset in training rather than the EgoVSR dataset corresponding to the test data, this further demonstrates the generalization ability of the motion blur synthesis model.

\subsection{Mask Loss}

\begin{figure}[h]
   \begin{center}
   \includegraphics[width=\linewidth]{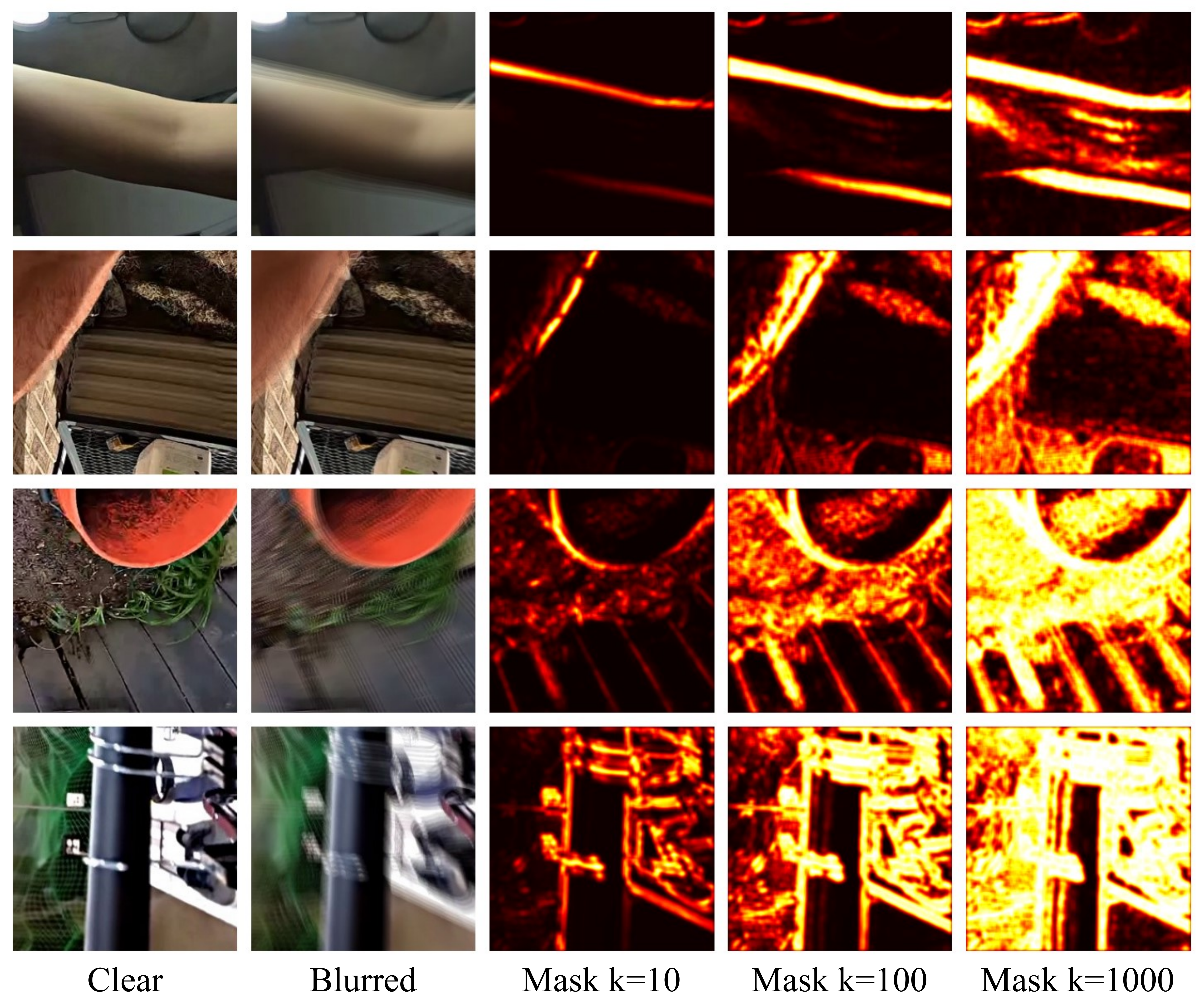}
   \end{center}
       
      \caption{Motion blur synthesis results and experiments with different magnification factors $k$.}
       
   \label{fig:blur-mask-exp}
\end{figure}

\begin{table}[h]
   \centering
   \caption{Explore the influence of magnification factor $k$.}
      \scalebox{1}{\begin{tabular}{cccc}
     \toprule
    $k$     & 10    & 100   & 1000 \\
     \midrule
     NIQE↓ & 5.0467 & \textbf{4.9678} &5.0804  \\

     PI↓   & 4.1756 & \textbf{4.0815} &4.1698   \\

     NRQM↑  & 6.6954 & \textbf{6.8047} &6.7407   \\
   \bottomrule
     \end{tabular}}%
      
   \label{tab:mask-k-exp}%
 \end{table}%

To demonstrate the effectiveness of our proposed blur synthesis model and mask ground-truth estimation method, we present the blur synthesis results and the estimated mask ground-truth under various magnification factors $k$ in Fig.~\ref{fig:blur-mask-exp}. The top two figures show the motion blur results from object motion, while the bottom two figures show the motion blur results from ego-motion. Our proposed model was able to synthesize both types of motion blur effectively.

Moreover, ground-truth of the mask has different sensitivity at different magnification factors $k$. 
When $k$ is small, the mask ground-truth is not enough to cover all the blurred areas.
As $k$ becomes larger, more noise areas are included in the mask ground-truth. 
The quantitative results on different magnification factors are shown in Tab.~\ref{tab:mask-k-exp}.
To ensure more balanced restoration results, we choose the magnification factor $k$=100 in our experiments.

\subsection{DB$^2$Net}
\begin{figure}[h]
   \begin{center}
   \includegraphics[width=\linewidth]{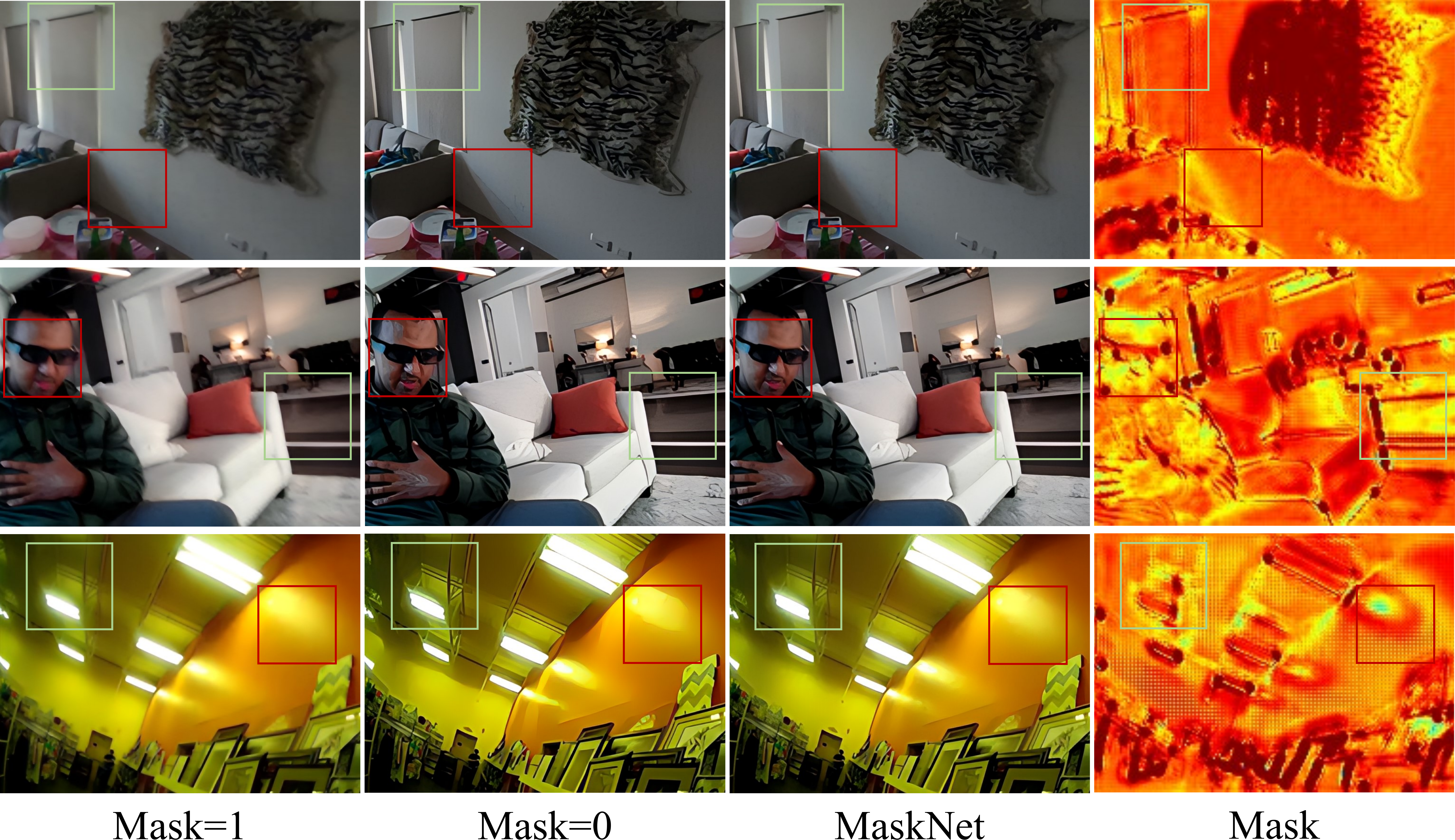}
   \end{center}
          
      \caption{Experimental results of different mask values. The green box indicates the area needs to be deblurred, and the red box indicates the over-sharpened area.}

   \label{fig:dual-branch-exp}
\end{figure}

In order to investigate the mechanism of the DB$^2$Net for deblurring, we use the same experimental setup as in Sec.~\ref{sec:setup} and manually set the blurring mask to all-0 and all-1. 
Since the inputs of the two branches are multiplied by $M$ and $1-M$, it will allow video frames to selectively pass through one of the branches by manually setting the blurring mask to 0 and 1. 
Results are shown in Fig.~\ref{fig:dual-branch-exp}.
We can observe that the network tends to restore a more diffuse and smoother result when the mask equals one and a sharper result when the mask equals zero. 
However, artifacts and oversharpening will appear in the face and shadow areas of the image when the mask equals zero. 
Conversely, the edges in an image are still blurred when the mask equals one.

Intuitively, a region with a mask $M$ value of 0 implies that the area is fully motion-blurred, and the blur branch needs to perform significant deblurring to ensure clear feature propagation. However, a full zero mask can lead to over-sharpening and undesired artifacts. In contrast, a mask value of 1 indicates a clear region, and the clear branch performs proper restoration without deblurring or sharpening.
Using the blurring mask estimated by MaskNet, our proposed DB$^2$Net can effectively combine the results of clear and blurred regions, enabling deblurring while reducing artifacts. This approach provides an effective way to balance the level of deblurring and sharpening needed for a given region, resulting in more natural-looking and visually pleasing results.
Overall, these observations help explain the mechanics of how our proposed DB$^2$Net works and highlight the importance of using a well-designed blurring mask to guide the restoration process for egocentric videos.

\subsection{Existence of clear frames}

\begin{figure}[h]
   \begin{center}
   \includegraphics[width=\linewidth]{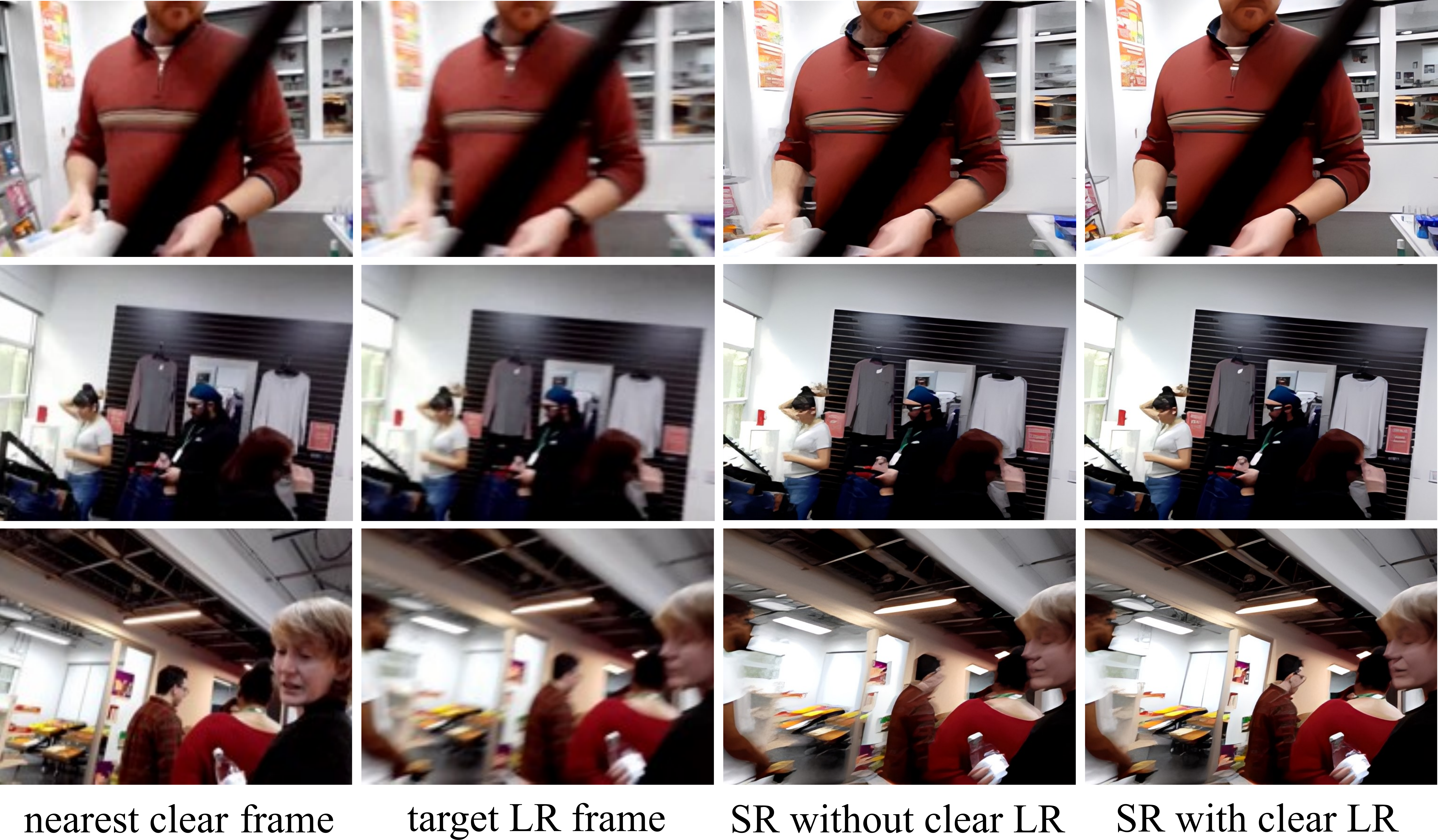}
   \end{center}
      \caption{Results of examples with and without clear frames. Clear frames exist within three frames around the target frames.}
   \label{fig:clear-frame-exp}
\end{figure}

The performance of VSR depends not only on the information of the current frame but also on the information of the past and future frames. 
To investigate the effect of the existence of clear frames on the deblurring and SR results, 
we filtered the clips with clear frames near the blurred target frames in the testset and replaced the clear frames with blurred frames to examine the clear frames' influence on the results.
Comparison results are shown in Fig.~\ref{fig:clear-frame-exp}.

When clear frames exist, the deblurring results are significantly better than the results when clear frames are excluded from the input data. 
The naturally existing clear frames in egocentric videos ensure that the network can propagate with real clear features, allowing the target frames to aggregate useful features from nearby frames. 
Experiments demonstrate that our EgoVSR indeed utilizes clear frames in multi-frame propagation and consequently obtains reliable egocentric VSR results.

\section{Discussion}

In this paper, we make the first attempt at egocentric video super-resolution research and propose a novel EgoVSR method that can restore egocentric videos both qualitatively and quantitatively. By synthesizing motion blur and designing a well-tailored architecture, our network can effectively address multiple degradations in egocentric videos while performing SR.
Despite our efforts to train the model using the EgoVSR dataset, the results were not as satisfactory as expected. We believe our model will achieve better performance if higher quality egocentric video datasets can be exploited.
Additionally, our method relies on a straightforward motion blur synthesis model, which may not accurately reflect the real motion blur of the extreme complexity. Further research in this area will also be conducive to boosting egocentric video super-resolution.
We hope that our work will inspire future research in egocentric video restoration.

\bibliographystyle{IEEEtran}
\bibliography{egbib}

\end{document}